\documentclass[letterpaper, 10 pt, conference]{ieeeconf}  

\IEEEoverridecommandlockouts                              

\usepackage{amsmath}
\usepackage{amssymb}
\usepackage{graphicx}
\usepackage{textcomp}
\usepackage{xcolor}
\usepackage{pgfplots}
\usepackage{subfigure}
\usepgfplotslibrary{groupplots}
\usepackage{tikz}
\usepackage{tikzscale}
\usepackage{hyperref}
\usepackage{cleveref}
\usepackage{float}
\usepackage{multirow}
\usepackage{algorithm}
\usepackage{algpseudocode} 
\usepackage{siunitx}
\usepackage{comment}
\usepackage{booktabs}
\usepackage{alphalph}

\crefname{figure}{Fig.}{Fig.}
\Crefname{figure}{Figure}{Figures}
\crefname{equation}{}{}
\Crefname{equation}{Equation}{Equations}

\newcommand{\ie}{\textit{i}.\textit{e}., }
\newcommand{\eg}{\textit{e}.\textit{g}., }

\title{\LARGE \bf
Robotic LEGO Assembly and Disassembly \\ from Human Demonstration
}

\author{Ruixuan Liu$^{1}$, Yifan Sun$^{2}$ and Changliu Liu$^{1}$ 
\thanks{*This work is in part supported by Siemens and MFI.}
\thanks{$^{1}$Ruixuan Liu and Changliu Liu are with Robotics Institute,
	Carnegie Mellon University,
	Pittsburgh, PA, 15213, USA.
        {\tt\small ruixuanl, cliu6@andrew.cmu.edu}%
}
\thanks{$^{2}$Yifan Sun is with the Carnegie Institute of Technology, Carnegie Mellon University, 5000 Forbes Avenue, Pittsburgh, PA, 15213, USA. {\tt\small yifansu2@andrew.cmu.edu}
}}

\begin{document}

\maketitle
\thispagestyle{empty}
\pagestyle{empty}

\begin{abstract}
This paper studies automatic prototyping using LEGO.
To satisfy individual needs and self-sustainability, this paper presents a framework that learns the assembly and disassembly sequences from human demonstrations. 
In addition, a digital twin is developed to verify the correctness of robot learning before deploying to the real world.
Moreover, an end-effector tool (EOT) is designed, which allows large industrial robots to easily manipulate LEGO bricks.
The proposed system is deployed to a FANUC LR-mate 200id/7L robot.
Experiments demonstrate that the proposed system can effectively learn the assembly and disassembly tasks from human demonstrations.
And the learned tasks are realized by the FANUC robot.
\end{abstract}

\section{Introduction}
There exist fast automatic prototyping methods for rigid parts (e.g., 3D printing) to meet customization needs \cite{10.1007/978-3-319-33609-1_2,review_additive,RASIYA20216896}. 
However, automatic prototyping for assembly remains challenging.
Moreover, due to environmental concerns and government regulations \cite{disassemble_article}, it is important that the prototyping system can achieve automatic assembly as well as disassembly, which is currently done by manual labor \cite{9817380}, to ensure sustainability.

\begin{figure*}
\subfigure[Assembly Alignment.]{\includegraphics[width=0.17\linewidth]{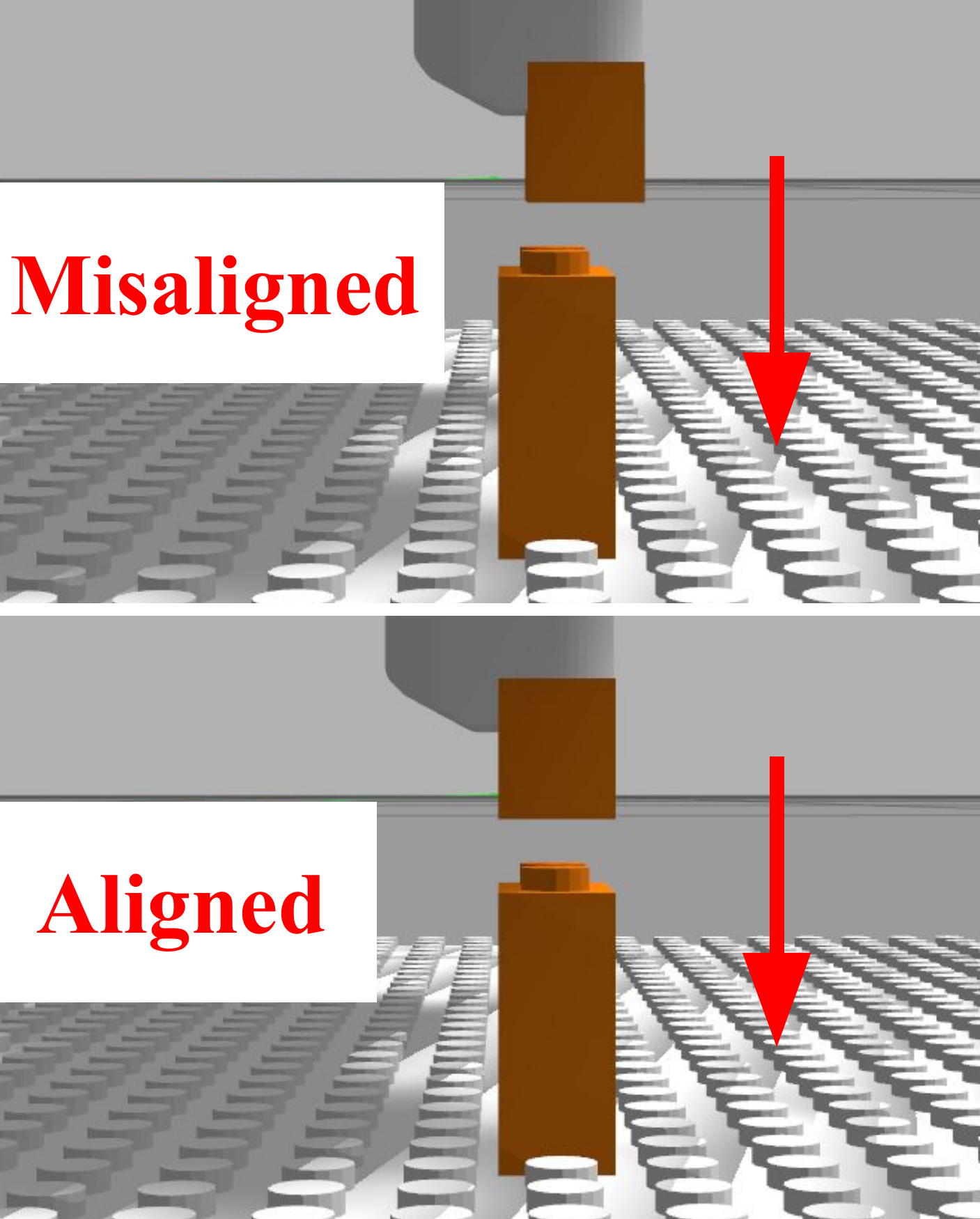}\label{fig:c1}}\hfill
\subfigure[Disassembly Criteria]{\includegraphics[width=0.17\linewidth]{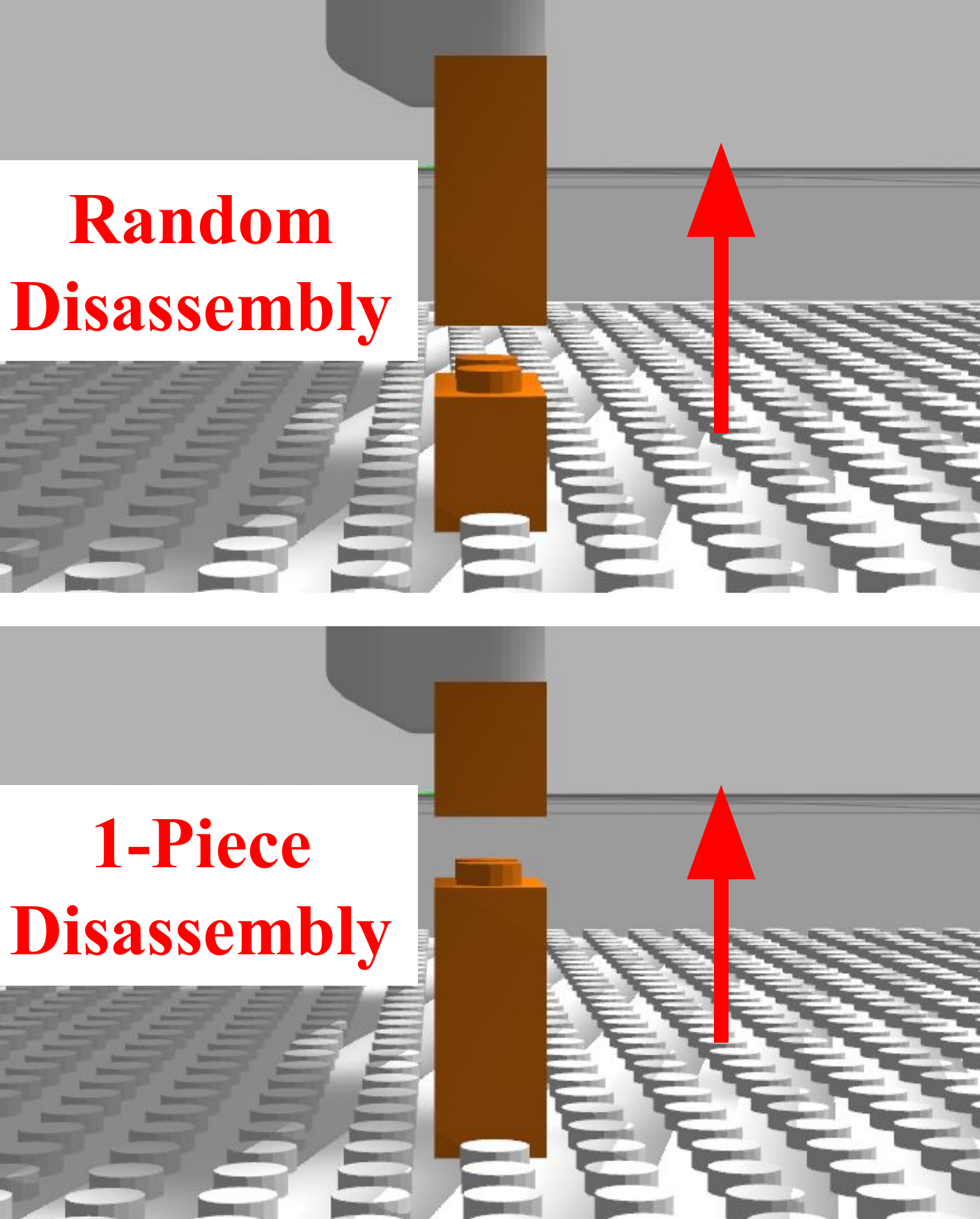}\label{fig:c2}}\hfill
\subfigure[User Preference]{\includegraphics[width=0.17\linewidth]{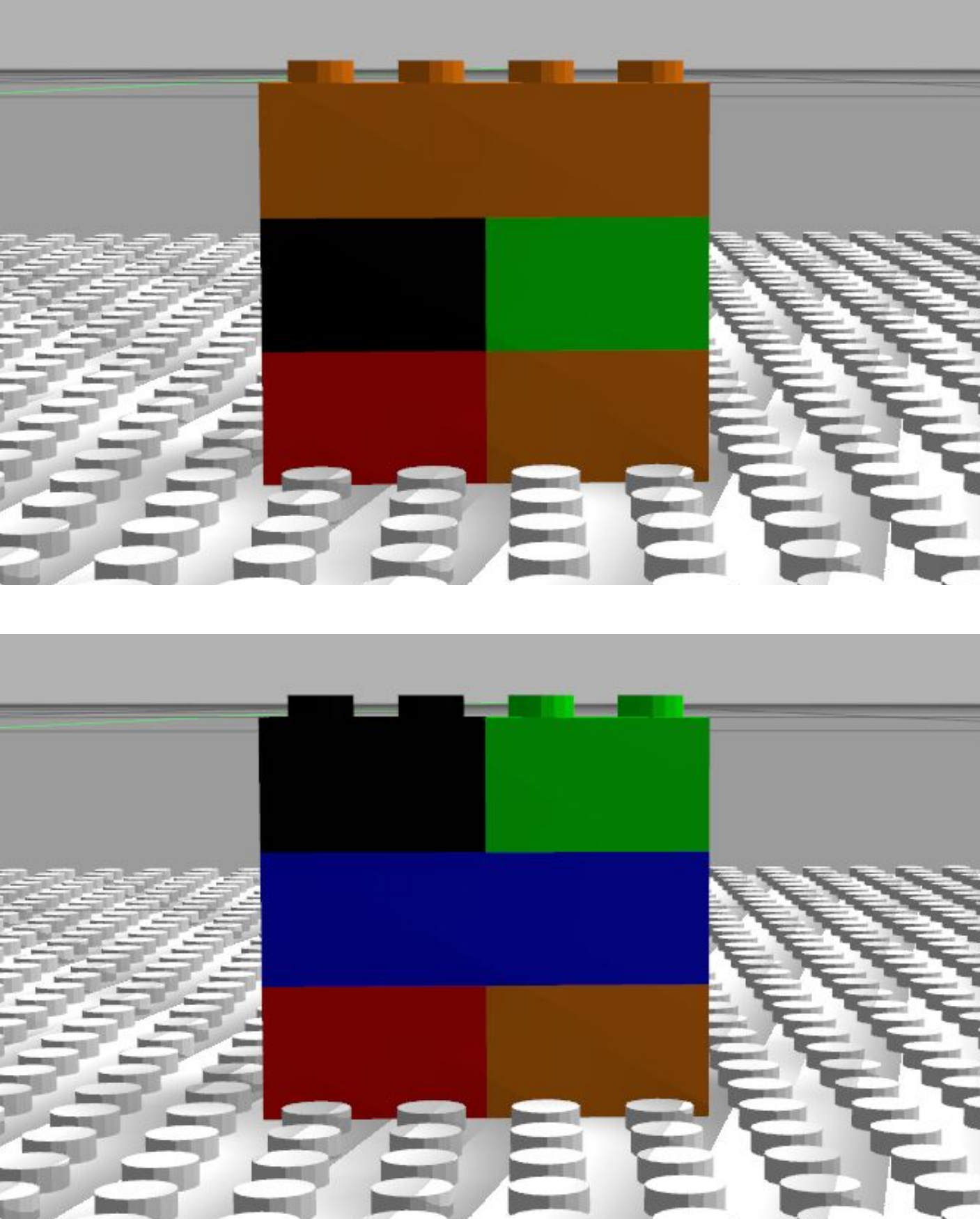}\label{fig:c3}}\hfill
\subfigure[Geometric Feasibility.]{\includegraphics[width=0.17\linewidth]{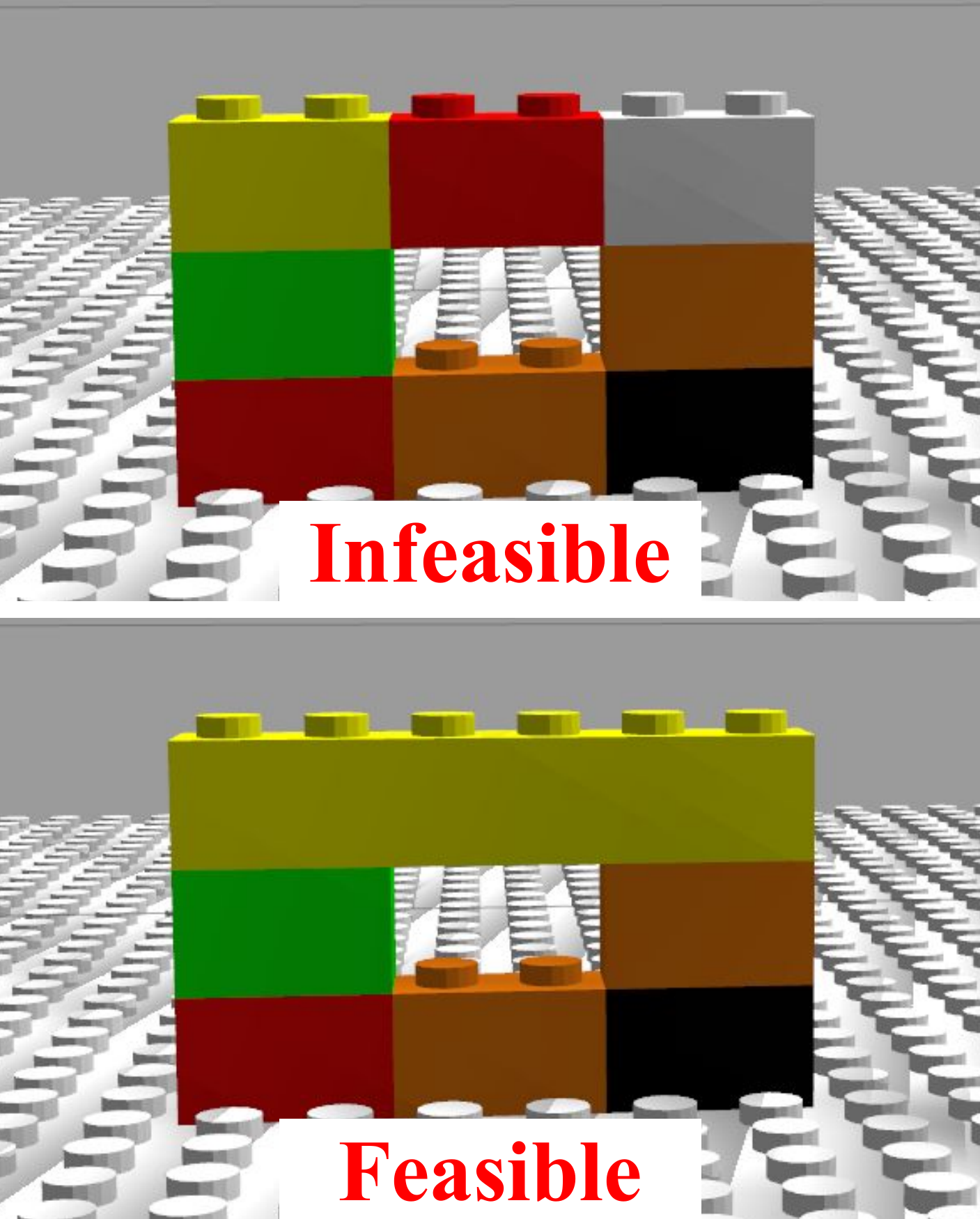}\label{fig:c4}}\hfill
\subfigure[Operability.]{\includegraphics[width=0.17\linewidth]{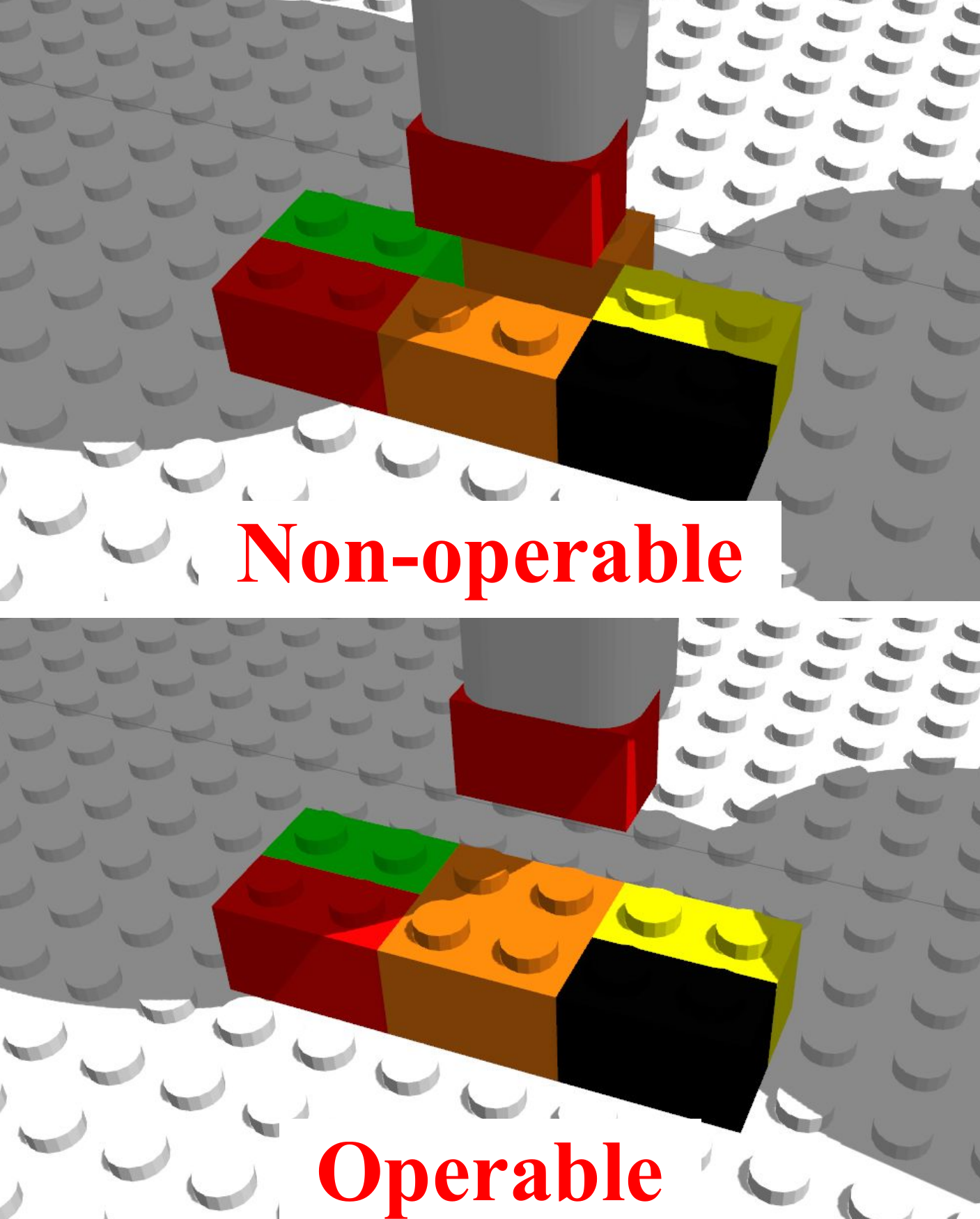}\label{fig:c5}}
\vspace{-5pt}
    \caption{\footnotesize Examples of requirements in LEGO assembly and disassembly. \label{fig:constraints}}
    \vspace{-10pt}
\end{figure*}

LEGO is a well-known platform for constructing pixelized objects.
In particular, this paper considers prototyping using LEGO.
\Cref{fig:constraints} shows examples of the major concerns in assembling and disassembling LEGO bricks, which make the problem non-trivial.
\Cref{fig:c1} shows the alignment constraint. Due to the tight fit of the LEGO bricks, two bricks should be aligned well in order to be assembled.
\Cref{fig:c2} shows the disassembly criteria due to different tightnesses between brick connections. 
It is not feasible to disassemble by directly pulling up the top brick since it would randomly drag up the bricks below it.
It is desired that one brick is disassembled at a time and the remaining structure is not broken.
\Cref{fig:c3} indicates the user preference. 
Different users are likely to have individual preferences (\eg colors, brick usage, etc).
It is important that the system can quickly accommodate the preferences.
\Cref{fig:c4} demonstrates the geometric constraint.
It is important that the bricks have no collision and that the LEGO structure is realizable in real.
The structure at the top is not feasible since there is no support under the middle brick (\ie red brick) in the top layer.
While the bottom structure is feasible since the top layer is switched to a longer brick, which is supported on both ends.
And lastly, \cref{fig:c5} illustrates the operability constraint.
To have a robot operating in the real world, it is important to ensure that there is enough space for the robot to manipulate the bricks and it has no collision with the environment.
Thus, the top image is not operable since the space is too tight.
It is desired that the middle brick is assembled first followed by the outside bricks, which is shown in the bottom picture.
In fact, LEGO assembly construction has been widely studied \cite{Kim2014SurveyOA,popov2017dataefficient, 7989334,10.1109/ICRA.2019.8793659,8674203,10.1145/2816795.2818091, doi:10.1177/09544054211053616,10.1111:cgf.13603,thompson2020LEGO,10.1007/978-3-031-19815-1_6,wang2022translating,LEGOBuilder,9341428,7759340,9812161}.
But the above-mentioned constraints remain challenging.

\begin{figure}
    \centering
    \includegraphics[width=0.9\linewidth]{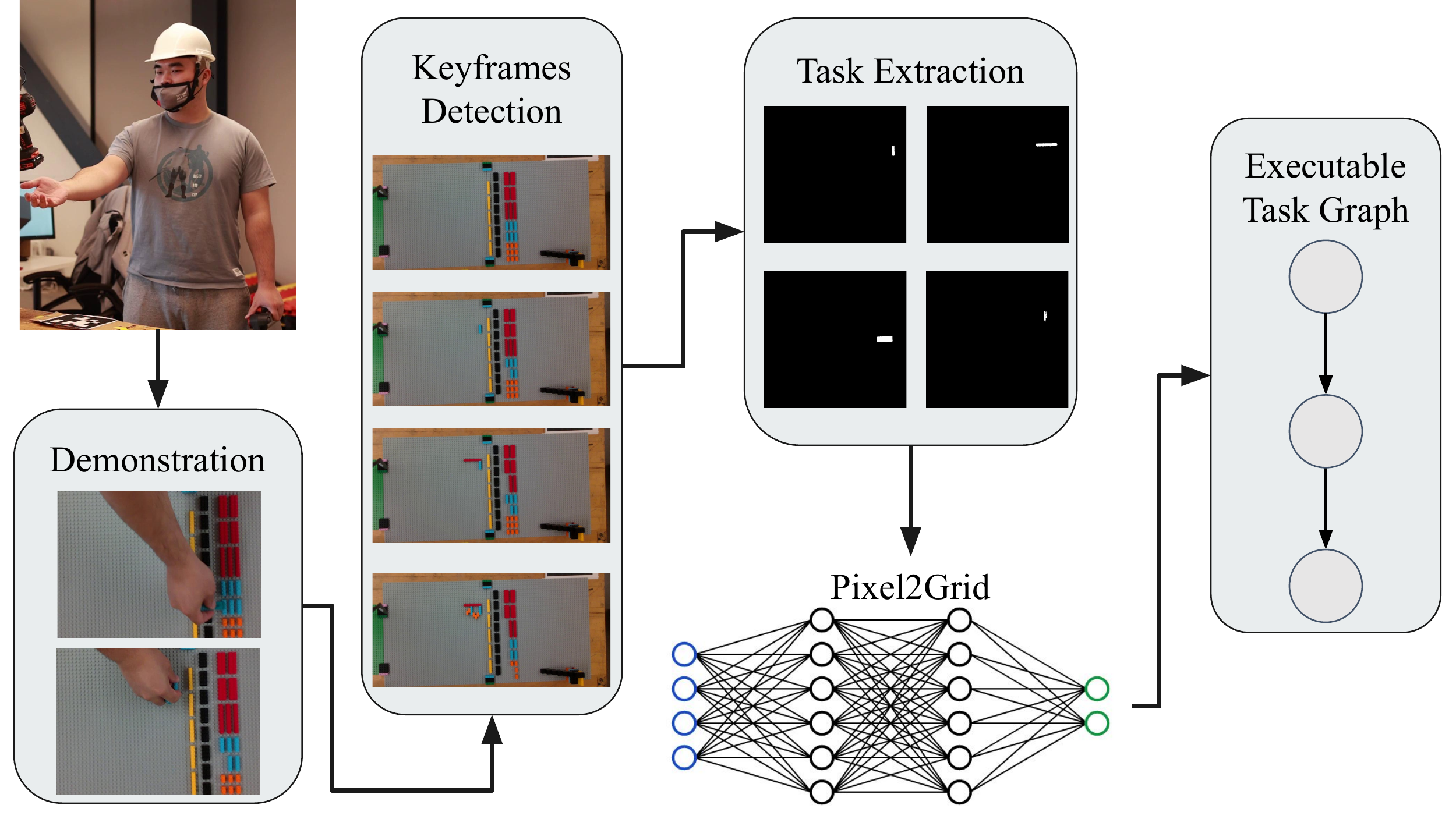}
    \caption{\footnotesize Illustration of the task learning from human demonstration. \label{fig:pipeline}}
    \vspace{-20pt}
\end{figure}


To address the challenges, this paper presents a fully automated pipeline that enables both robotic assembly and disassembly with LEGO.
To be specific, the system fulfills the user preference and geometric constraint by directly learning the task from human demonstration.
A digital twin is developed in order to verify the correctness of the learned task.
The task is verified if the operability is satisfied.
An end-effector tool (EOT) is designed for physical LEGO assembly and disassembly.
We deploy the system to a FANUC LR-mate 200id/7L robot. 
The experiments demonstrate that our system can effectively learn the assembly and disassembly tasks from human demonstrations.
The designed EOT allows the robot to assemble and disassemble LEGO objects reliably following the learned task knowledge.

\section{Method}
This paper presents a system that automates LEGO assembly and disassembly by learning from human demonstration.

\begin{figure}
\subfigure[\footnotesize EOT Design.]{\includegraphics[width=0.35\linewidth]{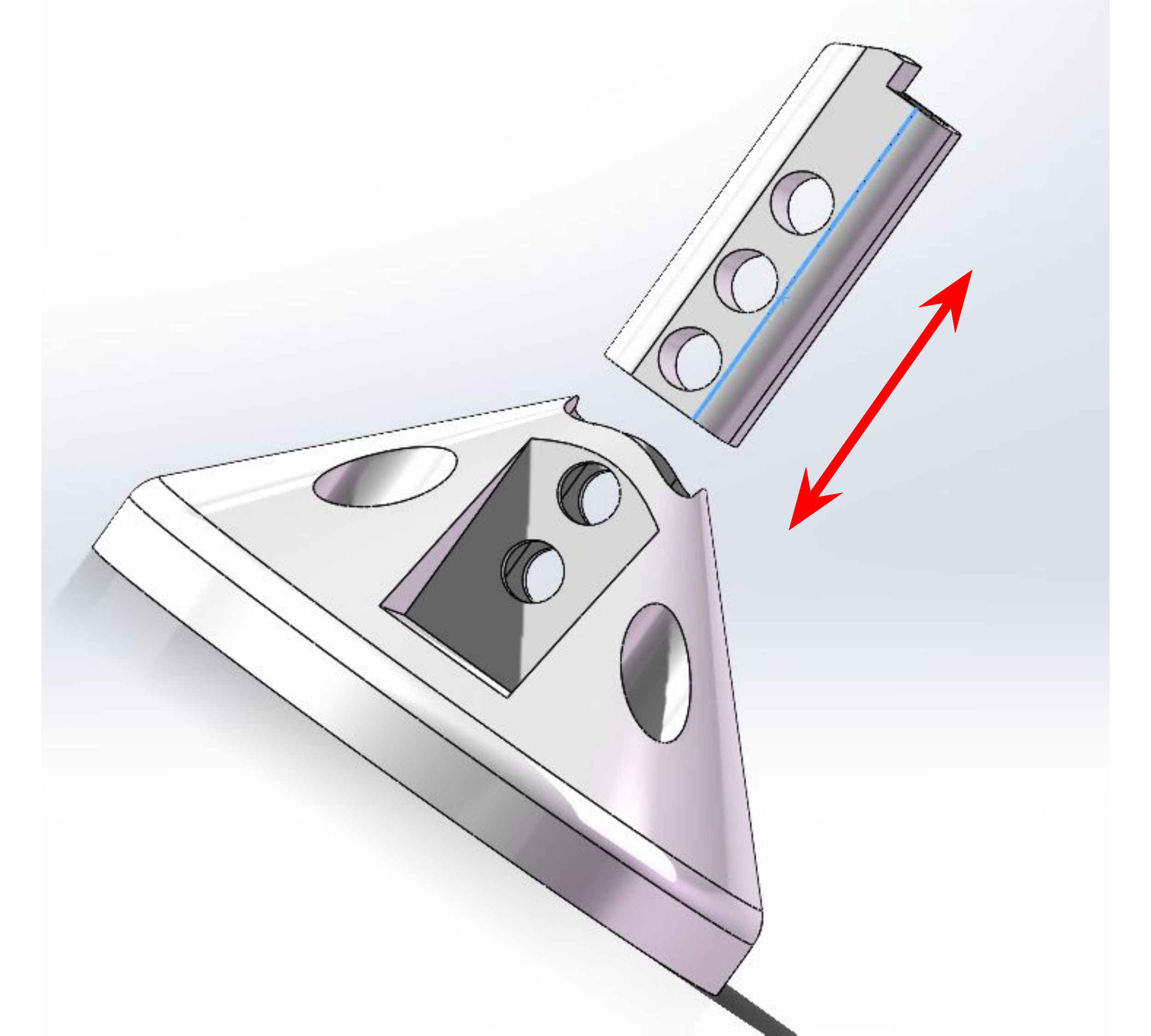}\label{fig:eot}}\hfill
\subfigure[\footnotesize Assembly and disassembly motion.]{\includegraphics[width=0.58\linewidth]{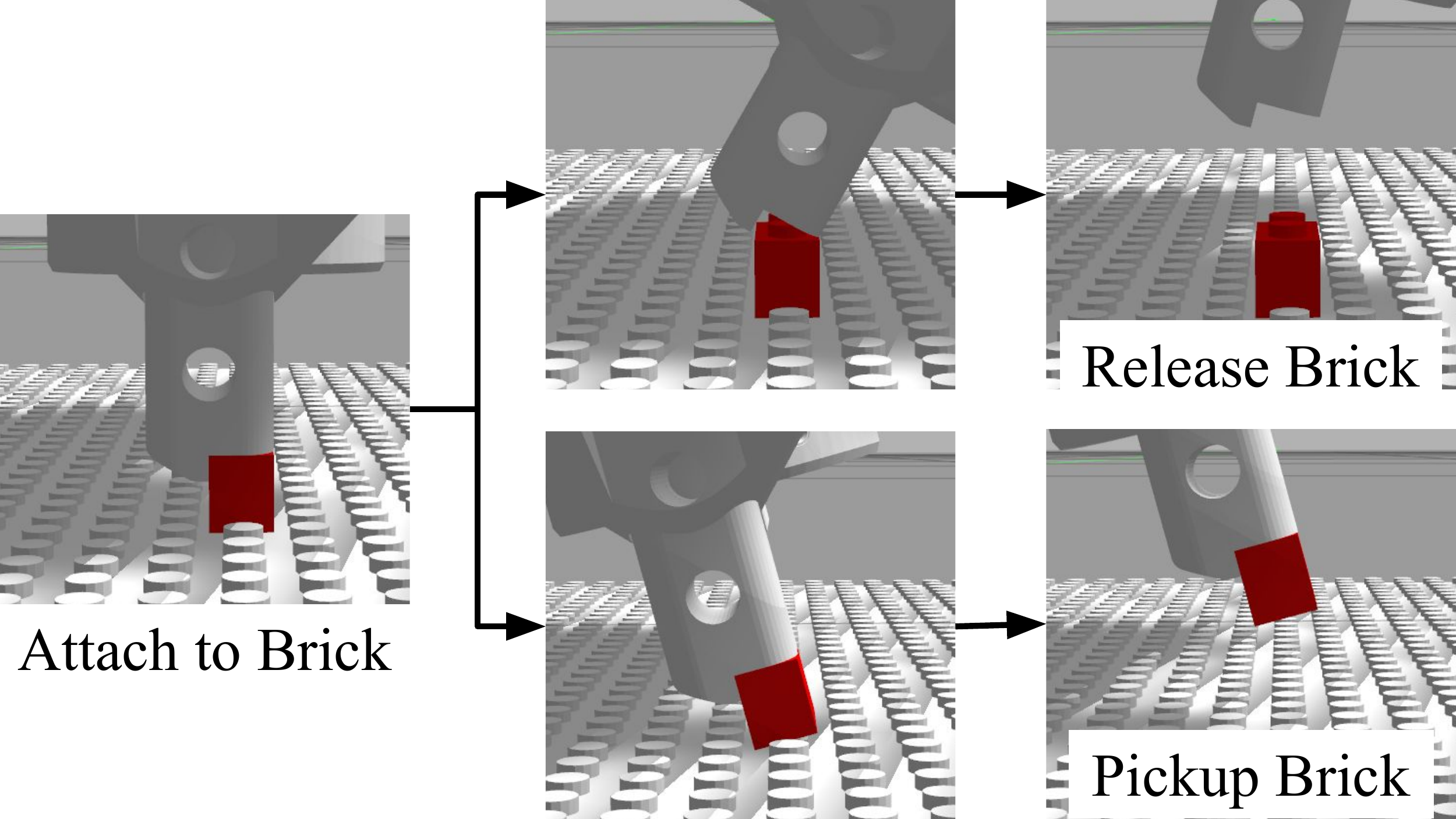}\label{fig:eot2}}
    \caption{\footnotesize LEGO Manipulation. \label{fig:manipulation}}
    \vspace{-20pt}
\end{figure}

\subsection{Task Learning from Human Demonstration}
\Cref{fig:pipeline} illustrates the learning framework.
The goal is to learn temporal task graphs that represent the assembly and disassembly sequences, which are denoted as $G=[g_1,g_2,\dots,g_T]$.
And $T$ is the task horizon.
Each node $g_i=\{id_i,~p^s_i,~\omega^s_i,~p^a_i,~\omega^a_i\}$ contains $id_i$, which defines the LEGO brick type (\ie shape and color). 
And $p^s_i$, $\omega^s_i$ and $p^a_i$, $\omega^a_i$ denote the brick position and orientation in the storage and assemble workspaces respectively.
Note that the learned task graph $G$ satisfies the user preference \cref{fig:c3} and geometric feasibility \cref{fig:c4} in nature.

As shown in \cref{fig:pipeline}, a human demonstrates LEGO assembly according to his preference. 
A depth camera (\eg Realsense) is used to capture the evolving workspace.
Then, the keyframes are detected, and the pipeline extracts the brick feature of the assembly step (\ie $id_i$). A pre-trained pixel-to-grid model generates the task information, \ie 1) where the brick is coming from (\ie $p^s_i,~\omega^s_i$) and 2) where the brick is assembled (\ie $p^a_i,~\omega^a_i$).
Note that the human only needs to demonstrate the assembly. 
The disassembly graph is constructed by reversing the assembly graph.

\subsection{Digital Twin for Task Verification}
\begin{figure}
 \centering
    \includegraphics[width=0.85\linewidth]{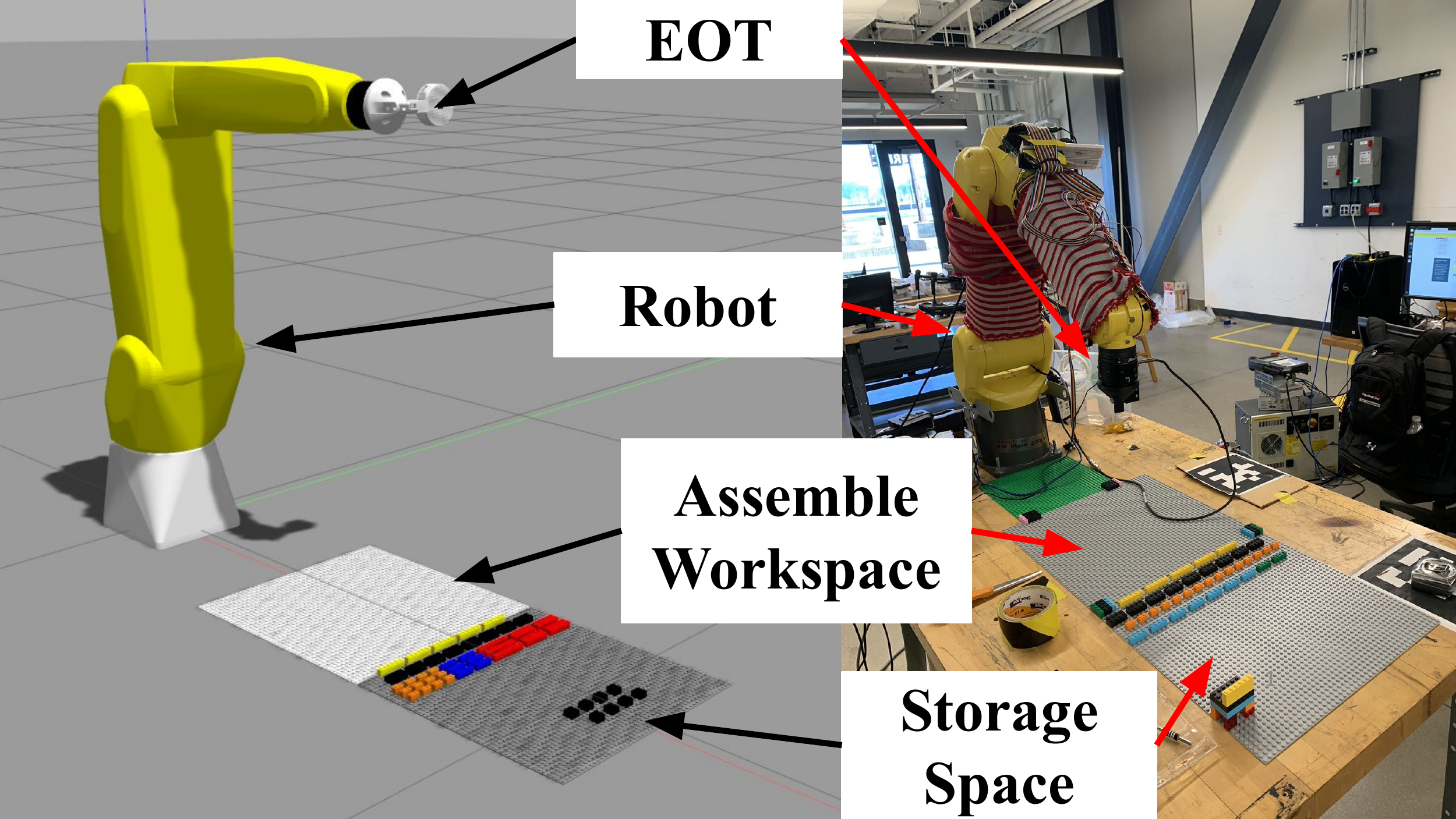}
    \caption{ Environment setup. Left: digital twin. Right: real setup.\label{fig:setup}}
    \vspace{-20pt}
\end{figure}

The learned $G$ might violate the operability constraint \cref{fig:c5} since humans are more flexible and dexterous than robots.
Thus, a digital twin is developed, as shown in \cref{fig:setup}, in order to verify the operability of $G$.
The digital twin shares the same robot control with the real system, and mimics the environment setup (\ie the robot model and LEGO bricks).
Given a $G$, the digital twin can execute it and check if it is operable (\eg collision-free, bricks are reachable, etc).

\subsection{EOT Design and LEGO Manipulation}
Manipulating LEGO bricks is challenging. 
This paper designs an EOT as shown in \cref{fig:eot}.
The EOT can be 3D printed. 
It consists of two main parts: 1) a base to mount to the robot and 2) a cylindrical tool, which is inspired by the LEGO separator \footnote{\url{https://www.lego.com/en-us/product/brick-separator-630}}. 
The length of the EOT is adjustable to satisfy different task requirements and this decoupled design allows the users to replace the worn parts without reprinting the whole EOT. 
As shown in \cref{fig:eot2}, LEGO manipulation can be completed with the designed EOT easily by attaching and twisting, which does not require any additional actuator.

\section{EXPERIMENTS}
To demonstrate our system, we deploy it to a FANUC LR-mate 200id/7L robot as shown in \cref{fig:setup}.
Two 48x48 LEGO plates are placed in front of the robot.
One is for storage and the other one is for assembly.
The assembly task is to use the available bricks in the storage space and realize the target LEGO object in the assemble workspace.
The disassembly task is to put back the bricks into the storage space.
The full experiment video is at \url{https://youtu.be/CRIKbV_0TvU}.


\section{CONCLUSIONS}
This paper presents a fully automated pipeline that enables both robotic assembly and disassembly with LEGO.
The system learns the assembly and disassembly tasks from human demonstration.
A digital twin is developed to verify the correctness of the learned task before deployment.
An EOT is designed for physical LEGO assembly and disassembly.
The system is deployed to a FANUC LR-mate 200id/7L robot and the experiments demonstrate that it can effectively learn the tasks from human demonstrations.
The designed EOT allows the robot to assemble and disassemble LEGO objects reliably following the learned task knowledge.

\bibliographystyle{ifacconf}
\bibliography{ifacconf}      
\end{document}